# PhysiNet: A Combination of Physics-based Model and Neural Network Model for Digital Twins

Chao Sun[1], Victor Guang Shi[1]

[1] *Advanced Manufacturing Research Centre, University of Sheffield, Rotherham, S60 5TZ, UK*

## Abstract

As the real-time digital counterpart of a physical system or process, digital twins are utilized for system simulation and optimization. Neural networks are one way to build a digital twins model by using data especially when a physics-based model is not accurate or even not available. However, for a newly designed system, it takes time to accumulate enough data for neural network models and only an approximate physics-based model is available. To take advantage of both models, this paper proposed a model that combines the physics-based model and the neural network model to improve the prediction accuracy for the whole life cycle of a system. The proposed hybrid model (PhysiNet) was able to automatically combine the models and boost their prediction performance. Experiments showed that the PhysiNet outperformed both the physics-based model and the neural network model.

## 1. Introduction

In the era of Industry 4.0, by serving as the real-time digital counterpart of a physical system or process, digital twin technology (Figure 1) is providing new methods for controlling the process, optimizing the system design and ultimately adding value to the physical part [1]. Usually, digital twins [2]–[4] were built by either physics-based modeling or data-driven method (Figure 2).

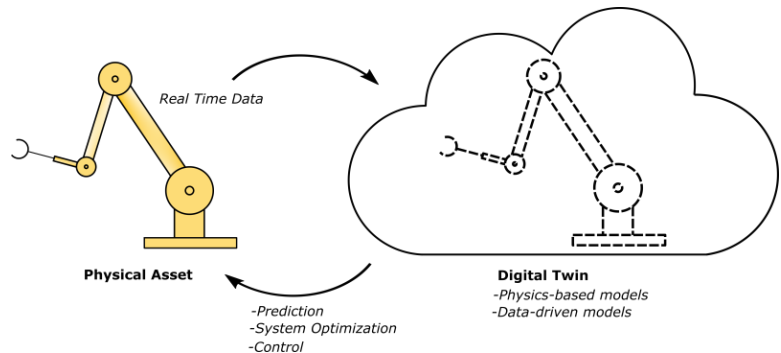

Figure 1. Digital twins for Industry 4.0.

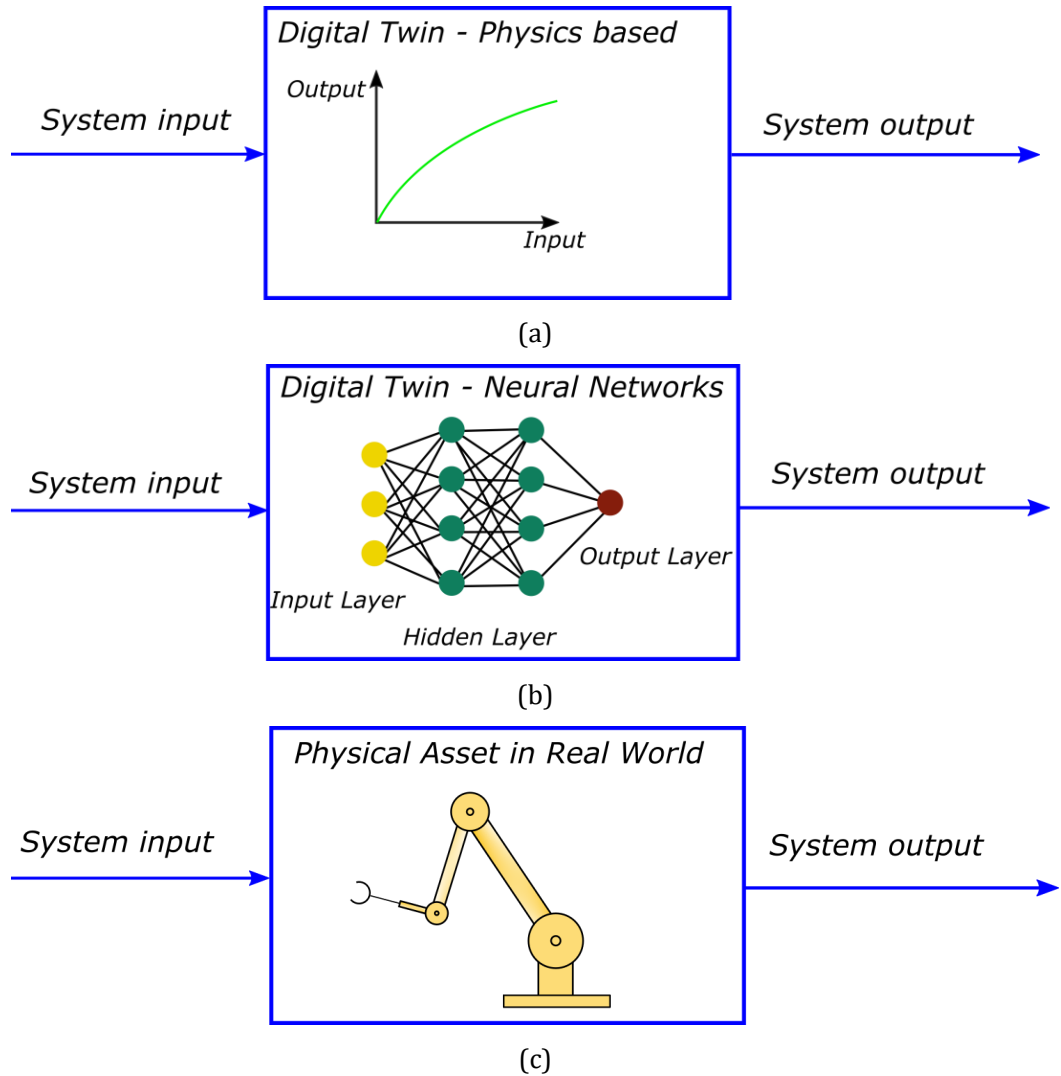

Figure 2. Comparison of digital twin models (a) Analytical model (b) Neural network model (c) Physical Asset.

Physics-based models were utilized to develop digital twin models for a variety of industry sectors, including aircraft systems [5], manufacturing process [6], machine tool systems [7] and robot systems [8]. Physics-based models require expert knowledge and advanced modeling techniques but they do not require any measurement data as input before the models start working. Figure 3 shows the architecture of a digital twin with a physics-based model. The model is developed based on physical laws and can start working right after it is developed. However, compared to data-driven models, physics-based models may be less accurate because they are sometimes over-simplified. For example, in many cases, to reduce the calculation complexity, the non-linearity of the system is ignored and the model is built with a linear equation. Moreover, parameters in physics-based models may not have been accurately measured or estimated. Those factors decrease the accuracy and fidelity of the physics-based digital twins. To improve the accuracy of a physics-based model, usually it requires engineers to redevelop the model by comparing the prediction and measurement (Figure 3).

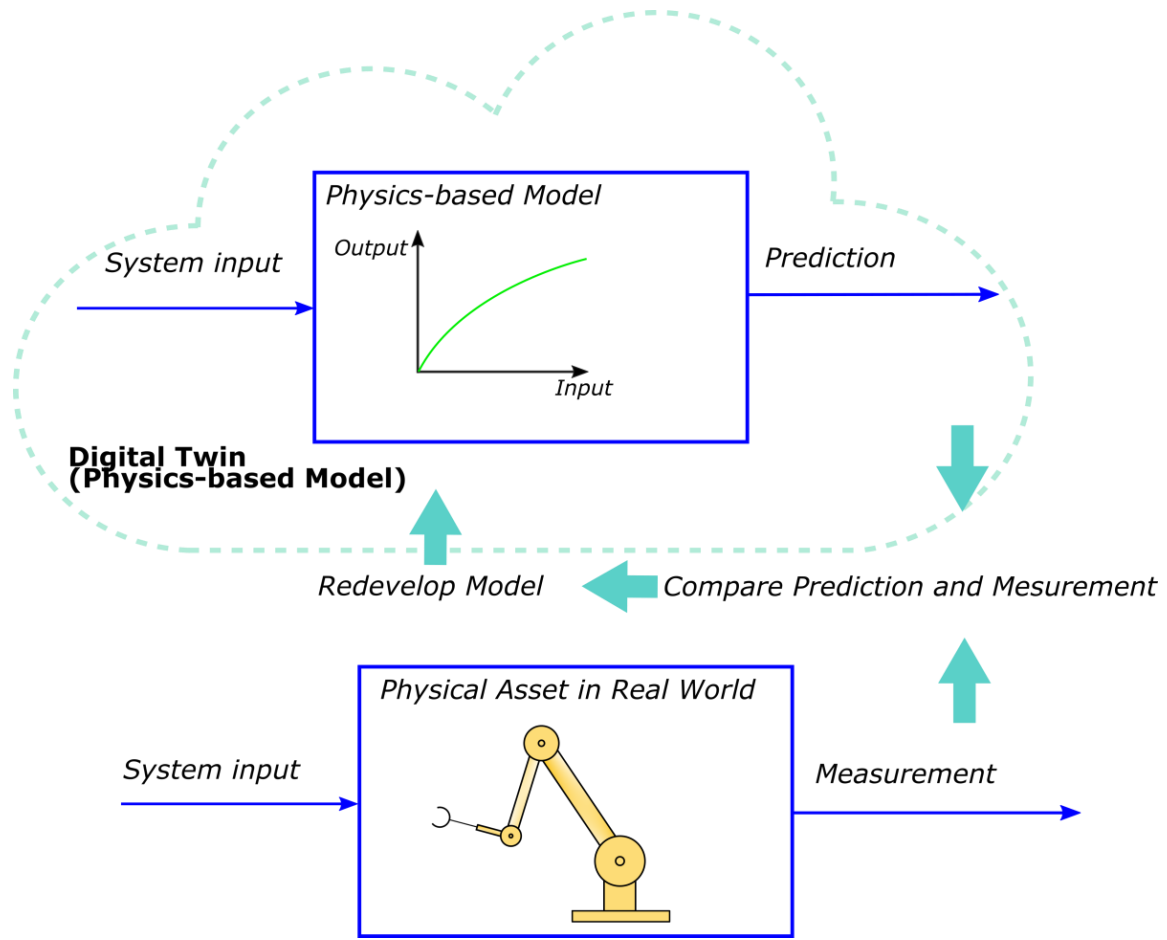

Figure 3. The architecture of a digital twin with a physics-based model.

Another method to build the digital twins is using data-driven methods. A machine-learning-based architecture for general sensor-fault detection, isolation, and accommodation (SFDIA) was proposed [9]. A neural network based method was developed for machining time prediction [10]. In these methods, the physical systems or processes were regarded as a black box and data-driven methods can be used to find the relationship between system input and output. Feed-forward neural networks are capable of approximating a function and its derivatives to arbitrary accurate [11]. Therefore, Neural networks were used for non-linear dynamic system modeling [12]. Figure 4 shows the architecture of a digital twin with a neural network model. Compared to physics-based models which require experienced engineers to do system identification and modeling, data-driven models do not need the user to fully understand the physical laws behind the system. As shown in Figure 4, to build the model, a neural network-based model only need the input and measurement from the system as the training set. However, the disadvantage of a data-driven model is that it requires a reasonable amount of data. If the quality and quantity of data are low, it is difficult to build a high-quality data-driven model. Therefore, usually, before a data-driven model start working properly, it needs to wait until the system accumulates enough data, especially when a new system or process is designed and developed and there is only a limited amount of data.

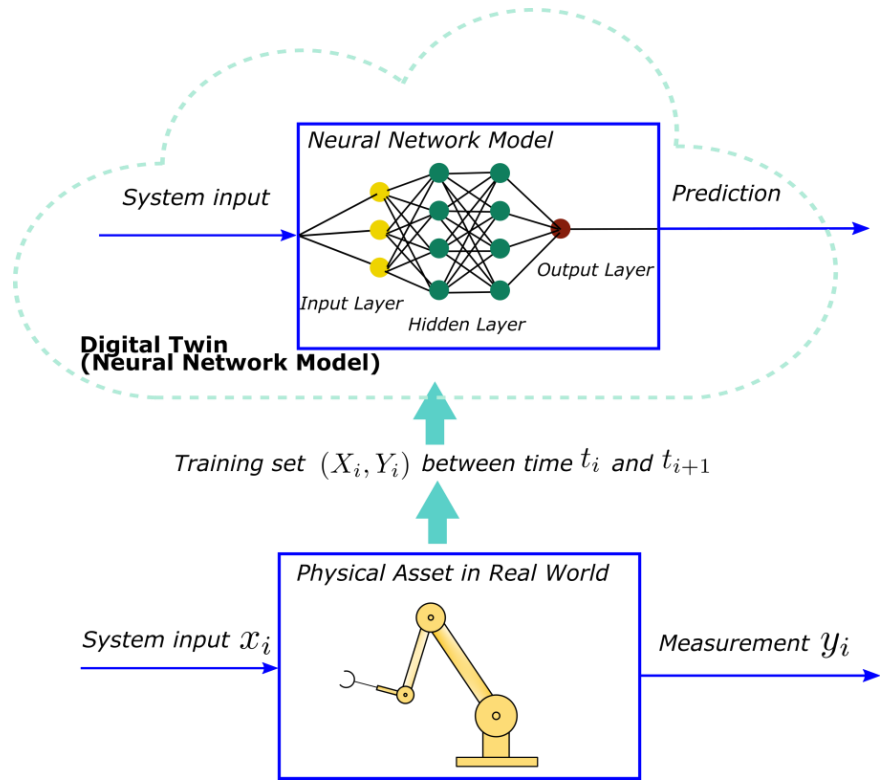

Figure 4. The architecture of digital twins with a neural network model.

To utilize the benefits of both the physics-based model and data-driven model, several methods have been developed. A physics-based model was used to generate data to train a data-driven model [13]. A physics-based model was also utilized to process data before a data-driven model[14]. Physics-informed deep neural networks were proposed [15]. It shows that introducing physics constraints increased the accuracy of neural network approximations. Conservative physics-informed neural networks were developed by substituting part of the data-driven model with a physics-based model [16].

Our main contributions are the following: a hybrid model (PhysiNet) which combines the physics-based model and the neural network model for digital twins was proposed. By using the information from both physics laws and data. Therefore, this digital twin model can serve the whole life cycle of a system (i.e. from the beginning when the data is not enough to build a proper data-driven model to the end when enough data was acquired).

The rest of this article is organized as follows. In Section 2, the structure and the mathematical model of this network are introduced. The physics-based model and neural network model are stacked into a network. By assigning different initial weights for each model, the combined model can utilize the benefits of both the physics-based model and the neural network model. In Section 3, two experimental cases are discussed to show the benefits of using the proposed model.

## 2. Method

The main idea behind PhysiNet is to combine the output from both the physics-based model and the neural network model by automatically updating weights for each of the models. Figure 5 shows the architecture of a digital twin with PhysiNet. The physics-based model and the neural network model are stacked into a network. For the time interval between $t_i$ and $t_{i+m}$, the system input $X = \{x_i, x_{i+1}, ..., x_{i+m}\}$ and measurement $Y = \{y_i, y_{i+1}, ..., y_{i+m}\}$ are collected and utilised as a training set $(X, Y)$ to train and update the digital twin (i.e. the PhysiNet).

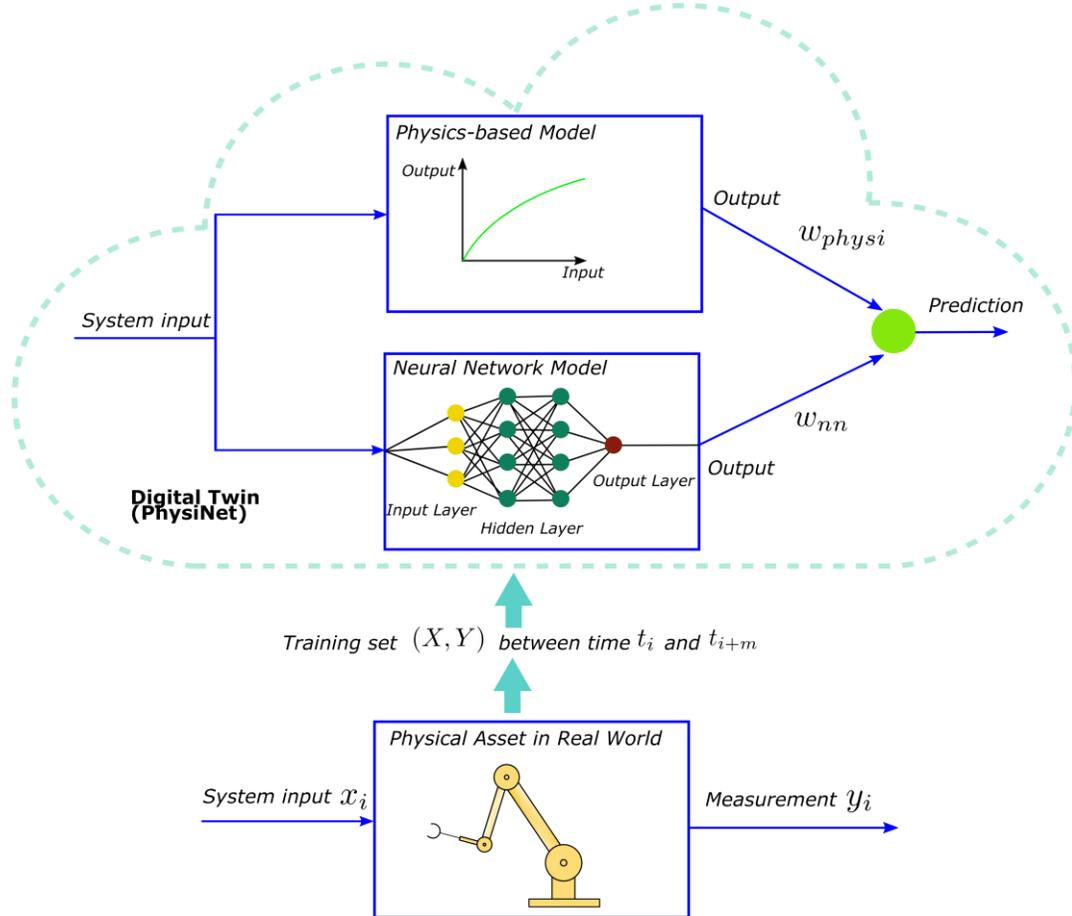

Figure 5. The architecture of a digital twin with PhysiNet.

Figure 6 shows the architecture of a PhysiNet with $n$ features $\{x_{i,1}, x_{i,2}, ..., x_{i,n}\}$ as the input, and one output. The features $\{x_{i,1}, x_{i,2}, ..., x_{i,n}\}$ are sent to both the physics-based part and the neural network part. From the physics-based part, $q_{physi}$ was calculated using function $f$. For neural network part, a feedforward neural network with $k$ hidden layers are used. The output of the $s^{th}$ neurons of the first hidden layer can be calculated by equation (1) where $active$ is the active function that introduces non-linearity to the neural network; $p_i$ is the input values to the network; $\omega_{k,s,i}$ is the weight parameters and $b_{k,s}$ is the bias parameters. If there are $m$ inputs coming from the last layer, the output of the $t^{th}$ neurons of the $k^{th}$ hidden layer can be calculated by equation (2). If there are $t$ inputs coming from last layer, the output of the $r^{th}$ neurons of the output layer can be calculated by equation (3), where *linear* is the linear transfer fuction. The final output of the neural network $q_{nn}$ can be calculated by equation (4). For the outputs $q_{physi}$ and $q_{nn}$, weights $w_{physi}$ and $w_{nn}$ are assigned to them respectively. The final output $\hat{y}_i$ can be calculated by equation (5). At the start of using PhysiNet, $w_{physi}$ and $w_{nn}$ are initialized by following the rule $\begin{cases} w_{physi} \gg w_{nn} \\ w_{physi} \to 1 \end{cases}$. For example, $w_{physi} = 0.99$ and $w_{nn} = 0.01$ (i.e. $\hat{y}_i \approx q_{physi}$). For the time interval between $t_i$ and $t_{i+m}$, the PhysiNet is trained with the training set $(X, Y)$ and weights in the neural network part, and weights $w_{physi}$ and $w_{nn}$ are all updated. what this procedure does is to find the best weights for each of the model to better fit the data set with both of the models.

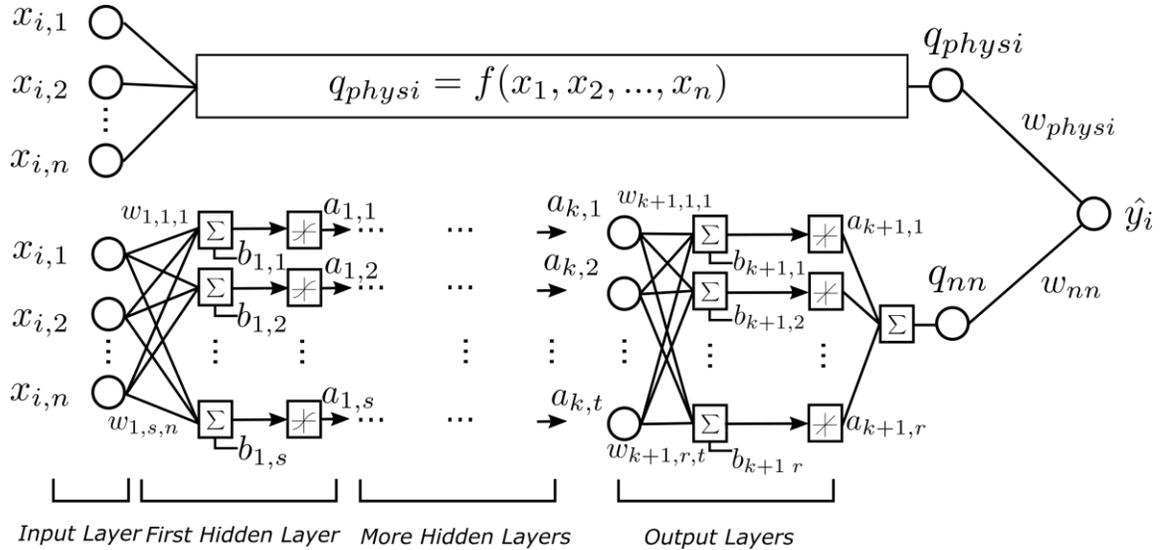

Figure 6. The architecture of PhysiNet.

$$a_{k,s} = active\left[\sum_{i=1}^{n}(p_i \cdot \omega_{k,s,i}) + b_{k,s}\right] \quad (1)$$

$$a_{k,t} = active\left[\sum_{i=1}^{m}(p_i \cdot \omega_{k,t,i}) + b_{k,t}\right] \quad (2)$$

$$a_{k+1,r} = linear\left[\sum_{i=1}^{t}(p_i \cdot \omega_{k+1,r,i}) + b_{k+1,r}\right] \quad (3)$$

$$q_{nn} = \sum_{i=1}^{r} a_{k+1,i} \tag{4}$$

$$\hat{y}_i = q_{physi} \cdot w_{physi} + q_{nn} \cdot w_{nn} \tag{5}$$

## 3. Experiments

### 3.1 Case study 1: polynomial equation

#### 3.1.1 Experimental Setup and Comparative Models

To validate the effectiveness of PhysiNet, a dataset was generated as the 'actual measurement data'. The measurement dataset was generated by $y_i = ax_i^2 + b + \sigma$, in which $a = 0.1$, $b = 15$ and $\sigma$ is the noise generated by a normal distribution with 0.5 standard deviation. The physics-based model was assumed to be a linear model defined by $q_{physi} = cx_i + d$, in which $c = 1$, $d = 10$. Figure 7 shows the measurement dataset and the physics-based model. The neural network part of PhysiNet was built with 2 hidden layers having 10 neurons. The weights for the neural network part were initialized randomly. $w_{physi}$ and $w_{nn}$ were initialized as 0.99 and 0.01 respectively.

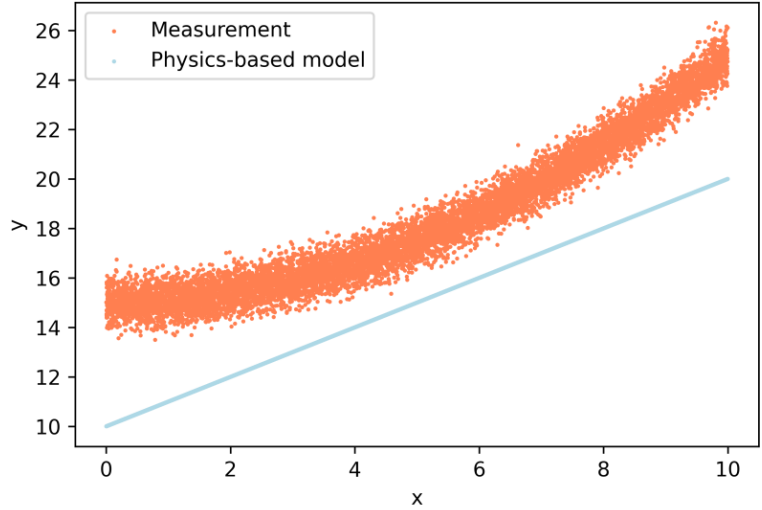

Figure 7. Physics-based model and measurement data.

To simulate the digital-twin process, for each 'time interval' (batch), 80 points were 'measured' (actually, generated) from the 'measurement data' for each step. In each step, the 80 points were used to train the model for 40 epochs. In total, there are 100 steps in the training. 2000 points from 'measurement data' were used as the test dataset. The performance was evaluated by mean squared error (MSE).

A digital-twin model with only a physics-based model and a digital-twin model with only a neural network model with 1 hidden layer having 4 neurons are also run on the dataset for comparison with PhysiNet.

### 3.1.2 Experimental Results

Table 1 gives the MSE results at step #0, #9, #19, #29, #39, #49, #59, #69, #79, #89 and #99. Compared to PhysiNet and the physics-based model, the neural network model had a larger MSE at the first step because the model was not trained yet. The physics-based model showed a constant MSE because it did not update throughout the steps. In between steps #0 to #9, the MSE of the neural network model and the MSE of PhysiNet became smaller than that of the physics-based model. From then, the MSE of PhysiNet kept the smallest among all three models. The digital-twin model with only a neural network model started having a close MSE with PhysiNet model in-between steps #39 to #49. Finally, at step #99, the MSE of the neural network model reached 0.377. At step #99, the PhysiNet model reached an MSE value of 0.278.

Table 1 MSE for PhysiNet, Neural Network Model and Physics-based Model.

| Step # | PhysiNet | Neural Network Model | Physics-based Model |
|--------|----------|----------------------|---------------------|
| 0      | 12.978   | 344.863              | 11.824              |
| 9      | 0.380    | 0.876                | 11.824              |
| 19     | 0.338    | 1.007                | 11.824              |
| 29     | 0.290    | 0.840                | 11.824              |
| 39     | 0.283    | 1.021                | 11.824              |
| 49     | 0.275    | 0.404                | 11.824              |
| 59     | 0.295    | 0.406                | 11.824              |
| 69     | 0.278    | 0.386                | 11.824              |
| 79     | 0.282    | 0.404                | 11.824              |
| 89     | 0.284    | 0.369                | 11.824              |
| 99     | 0.278    | 0.377                | 11.824              |

Figure 8 shows the change of weight ratio ($w_{physi}/w_{nn}$) during the steps. The ratio dropped below 2 within 20 steps from a value of 99, which means the neural network part of PhysiNet gradually start taking an important role in the PhysiNet model. Finally, the ratio went to a value close to 1, which means both the physics-based part and the neural network part of PhysiNet influenced the prediction results from the PhysiNet model in the end.

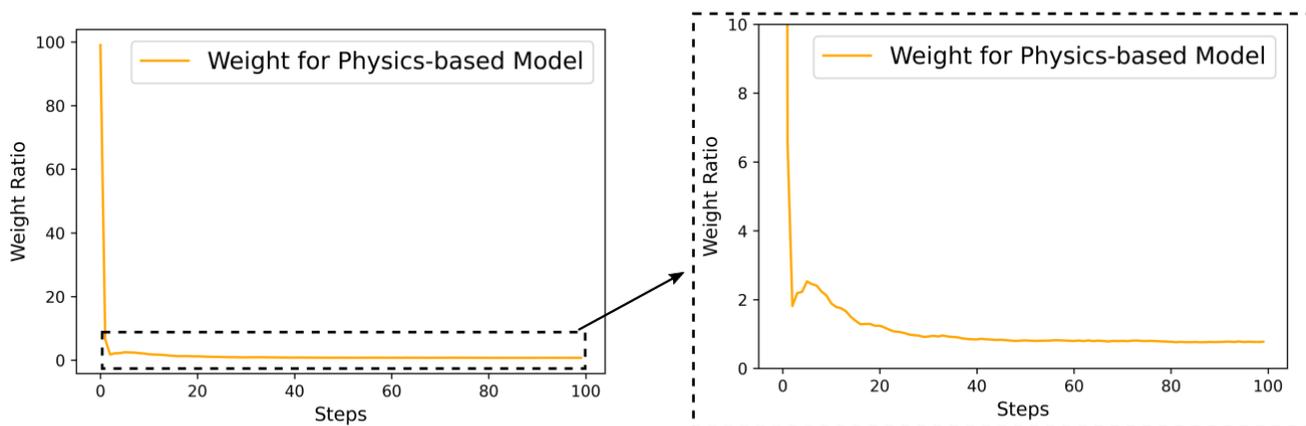

Figure 8. Change of Weight Ratio.

Figure 9 shows the prediction results from PhysiNet at steps #0, #9, #19, #29, #39 and #49. At step #0 and step #9, the PhysiNet model was mostly dominated by the linear physics-based part. At steps #19, #29, #39 and #49, the PhysiNet model learned the non-linear behavior of the system because the neural network part started having an important influence on the prediction.

Figure 10 shows the prediction results from the neural network model at steps #0, #9, #19, #29, #39 and #49. The neural network gradually improved the fitting. Starting from #49, the neural network model captured the non-linear behavior of the system.

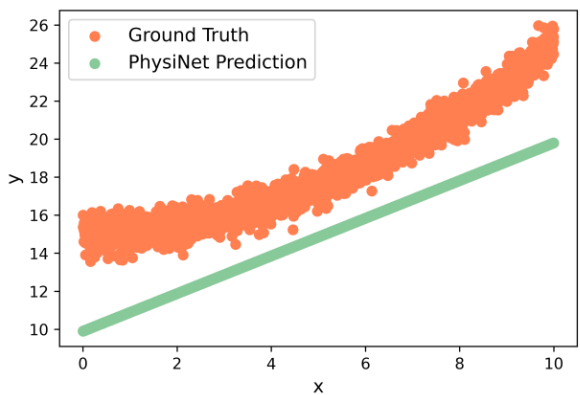
(a)

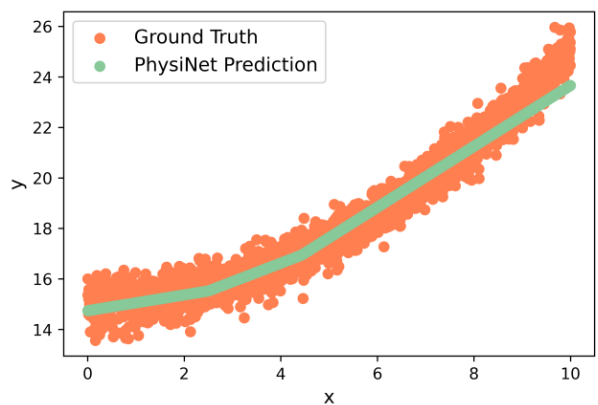
(b)

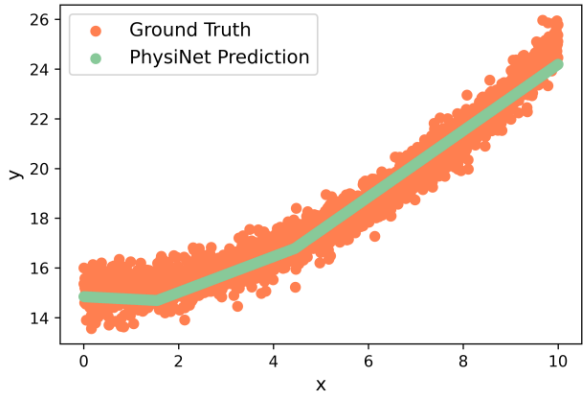

(c)

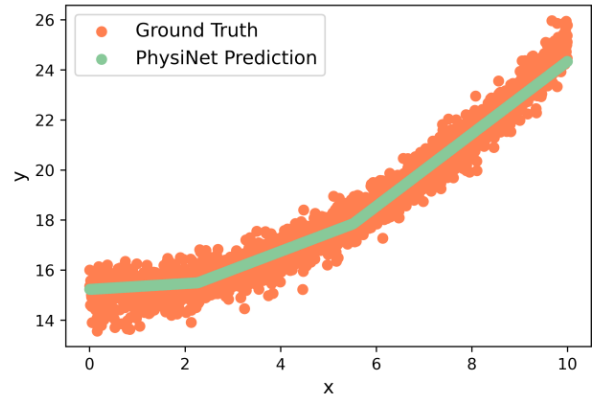

(d)

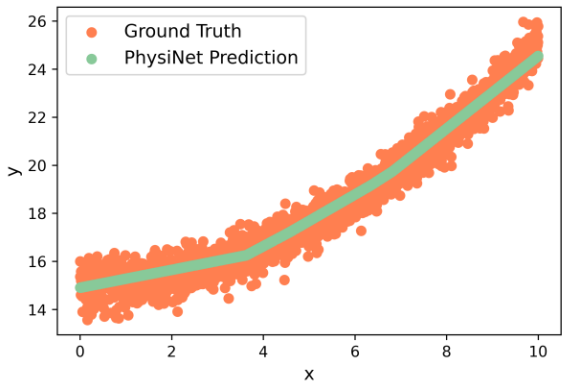

(e)

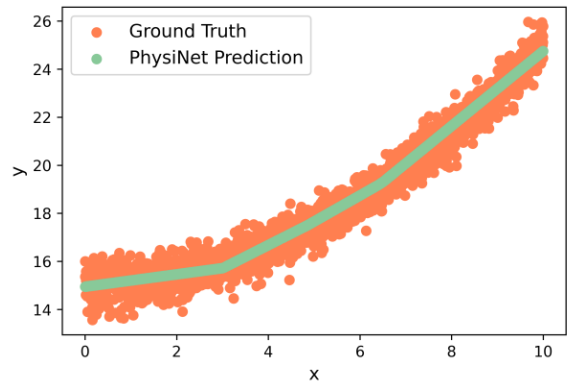

(f)

Figure 9. Progress of prediction from Physinet at step (a)#0(b) #9 (c) #19 (d) #29 (e) #39 (f) #49

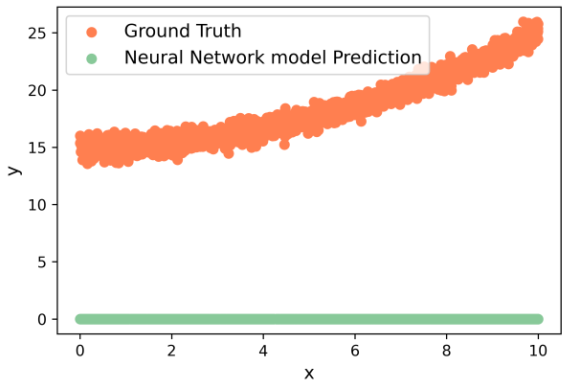

(a)

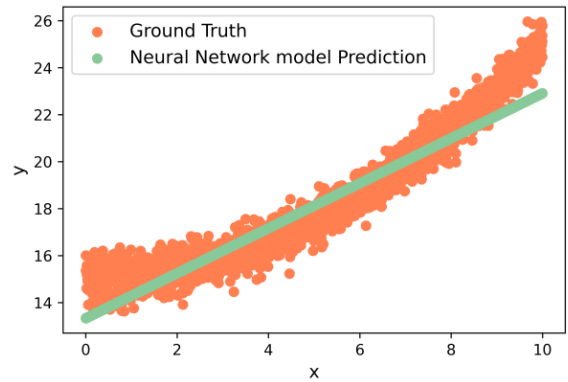

(b)

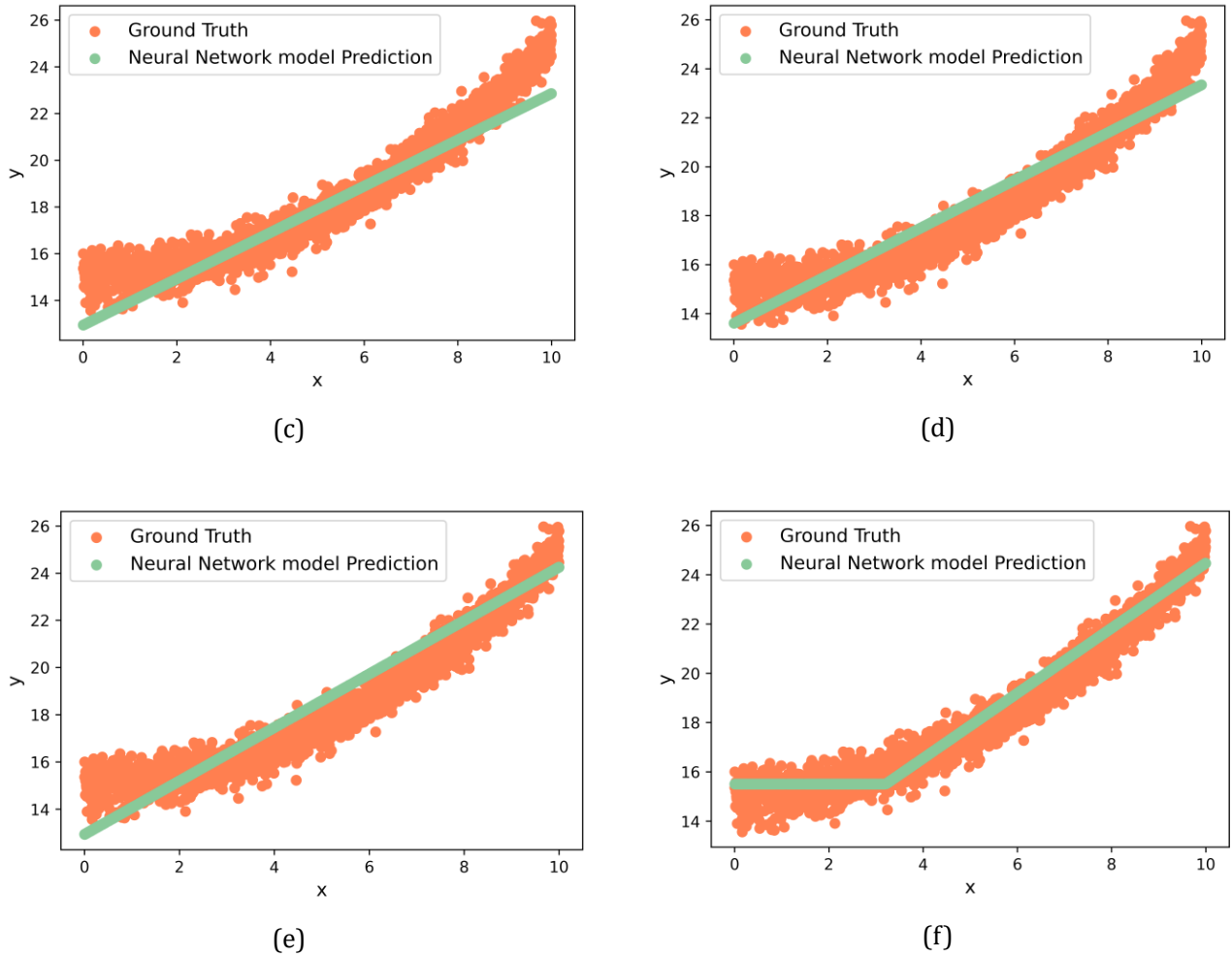

Figure 10. Progress of prediction Neural Network model at step (a) #0(b) #9 (c) #19 (d) #29 (e) #39 (f) #49

## *3.2 Case study 2: digital twin for a control system in the frequency domain*

### 3.2.1 Experimental Setup and Comparative Models

Case study 2 is to simulate and demonstrate how this model can be used for a real digital twin scenario. Figure 11 shows the diagram of a system in the Laplace domain for the ground truth and Figure 12 shows the diagram of the physics model in the Laplace domain. The system is composed of a gain, a controller and a plant. For the real system, the transfer function of the plant is $\frac{1}{s^2+s+4.1}$ and we assume that the transfer function of the model is $\frac{1}{s^2+s+4.4}$ because of the inaccuracy during the modelling process. Figure 13 shows the frequency response function for the real system and the physics-based model. It is assumed that for each time the system was used, we'll be able to measure 80 points of data to build the model. Each time the points were used to train the model for 40 epochs. In total, there are 100 steps in the training. 2000 points from the 'measurement were used as the test dataset. The performance was evaluated by mean squared error (MSE).

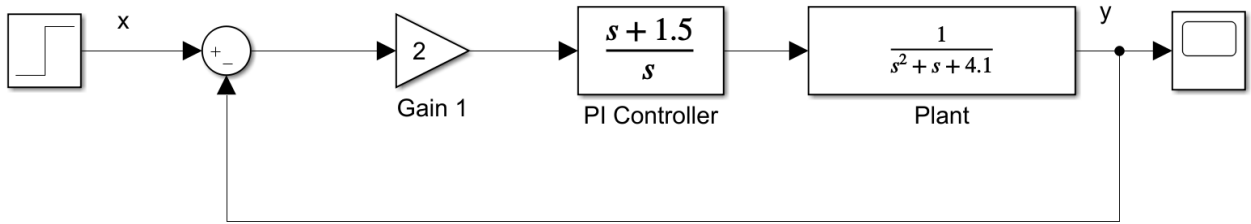

Figure 11. Diagram for the real system.

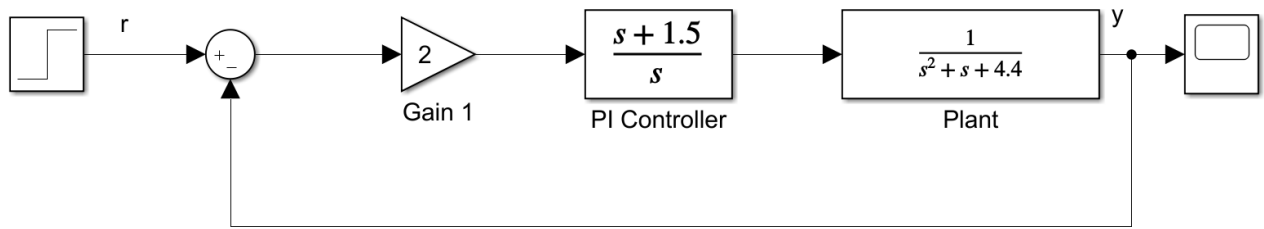

Figure 12. Diagram for the physics-based model.

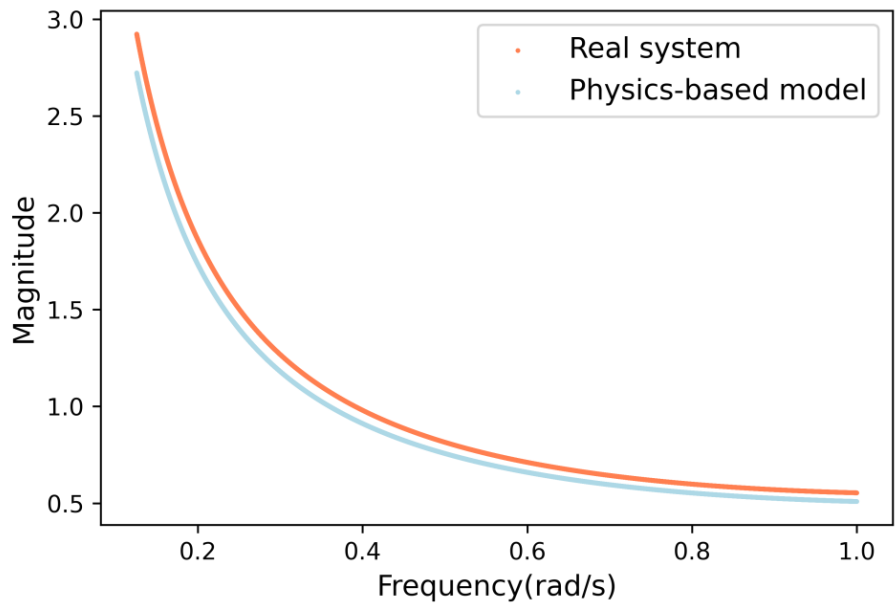

Figure 13. Frequency response function for the Physics-based model and the real system.

### 3.2.2 Experimental Results

Table 2 gives the MSE results at steps #0, #9, #19, #29, #39, #49, #59, #69, #79, #89 and #99. The neural network model had a larger MSE at the first step. Between steps #0 to #9, the MSE of the neural network model and the MSE of PhysiNet became smaller than that of the physics-based model. From then, the MSE of PhysiNet kept the smallest among all three models. Finally, at step #99, the MSE of the neural network model reached 5.070E-06. At step #99, the PhysiNet model reached an MSE value of 1.082E-03.

Table 2 MSE for PhysiNet, Neural Network Model and Physics-based Model.

| Step # | PhysiNet | Neural Network Model | Physics-based Model |
| --- | --- | --- | --- |
| 0 | 1.338E-02 | 2.171E+00 | 1.040E-02 |
| 9 | 1.360E-06 | 2.300E-03 | 1.040E-02 |
| 19 | 1.238E-06 | 1.544E-03 | 1.040E-02 |
| 29 | 4.417E-06 | 1.230E-03 | 1.040E-02 |
| 39 | 3.158E-06 | 1.034E-03 | 1.040E-02 |
| 49 | 4.329E-06 | 1.403E-03 | 1.040E-02 |
| 59 | 1.237E-06 | 1.113E-03 | 1.040E-02 |
| 69 | 1.591E-06 | 1.140E-03 | 1.040E-02 |
| 79 | 1.919E-06 | 1.102E-03 | 1.040E-02 |
| 89 | 1.517E-06 | 1.305E-03 | 1.040E-02 |
| 99 | 5.070E-06 | 1.082E-03 | 1.040E-02 |

Figure 14 shows the prediction results from PhysiNet at steps #0, #9, #19, #29. At step #0, the PhysiNet model was mostly dominated by the linear physics-based part. Since step #9, the PhysiNet model learned the behavior of the system because the neural network part started influencing the prediction.

Figure 15 shows the prediction results from the model with only neural network at steps #0, #9, #19, #29. Apart from #0, the neural network model captured the non-linear behavior of the system.

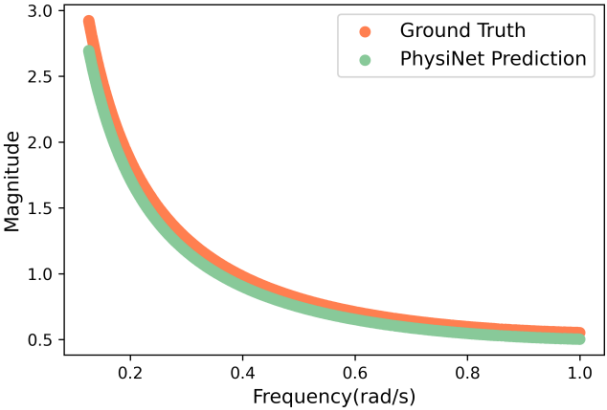
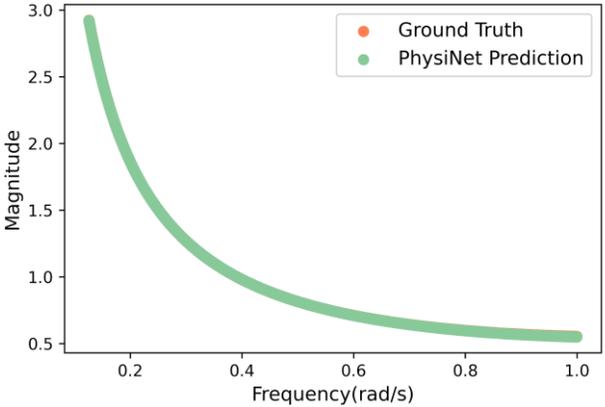

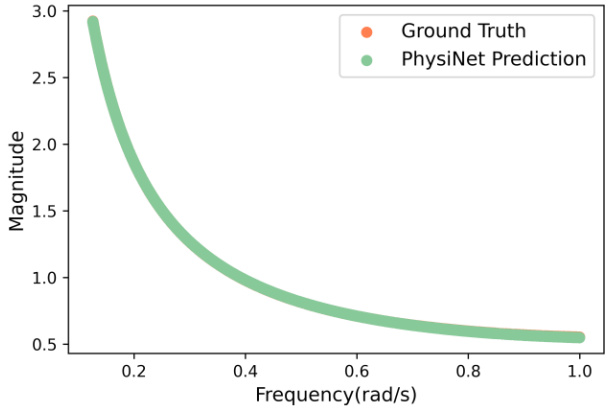
(a)

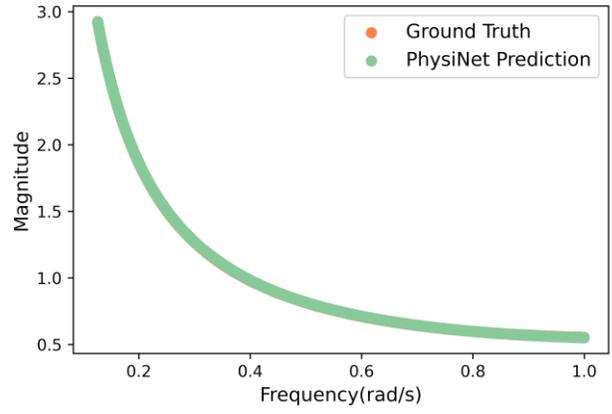
(b)

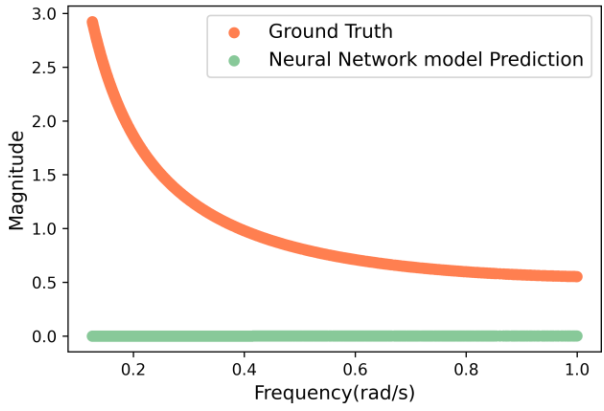
(c)

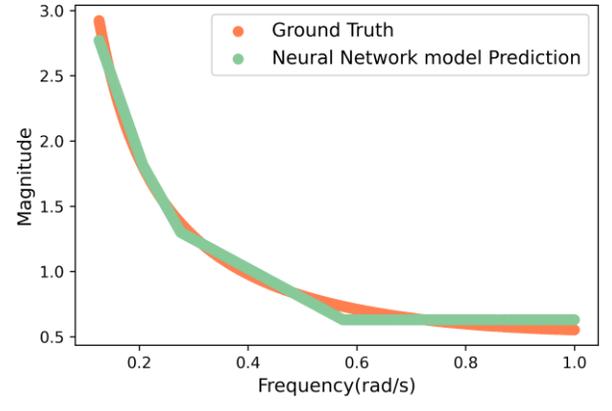
(d)

Figure 14. Progress of prediction from Physinet at step (a)#0(b) #9 (c) #19 (d) #29

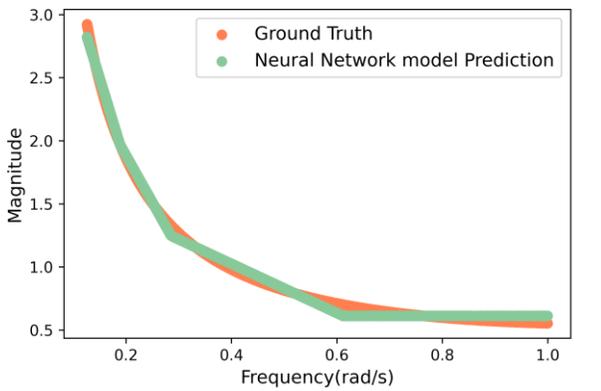
(a)

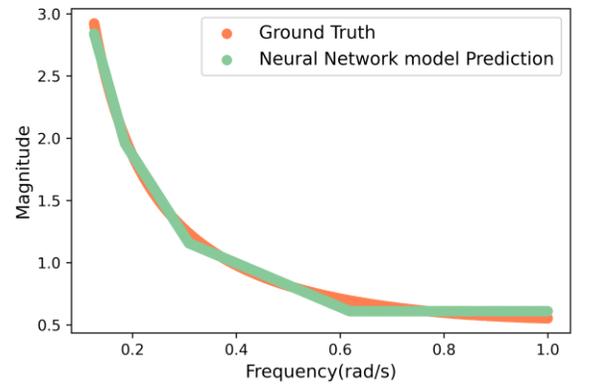
(b)

(c)

(d)

Figure 15. Progress of prediction Neural Network model at step (a)#0(b) #9 (c) #19 (d) #29

## 4. Conclusion and Future Work

This paper has presented a novel framework for the building of digital twin models. The main findings are as follows:

1. The research presented a digital twin model combining both physics-based model and neural network model by automatically assigning different weights to each of the models. The model was designed to serve as a digital twin model for real machines or processes for their whole life cycle, including both the beginning when there is not enough data and the end when neural network solution can provide a model with higher accuracy. Simulation results in this paper showed that the PhysiNet model gave lower MSE than the neural network model at the beginning and lower MSE than the physics-based model after the beginning of the steps.

2. In the simulation, It is observed that both PhysiNet and neural network model reached an SME lower than that of the physics-based model. However, it was also seen that finally, the PhysiNet reached a lower MSE than that of the neural network model. The reason could be that adding a physics-based model helped PhysiNet find better local minima. This phenomenon needs further investigation in the future.

3. In this paper, the method was developed for regression problems in developing a digital twin. It is worth a further investigation to explore the possibility of using this method for other problems, like computer vision, in digital twins in the future.

## Acknowledgment

The authors are grateful for the support of the Engineering and Physical Science Research Council through the Digitally Enhanced Advanced Services NetworkPlus funded by grant ref EP/R044937/1 and STFC Food Network RC grant reference ST/V001450/1.